\documentclass{article}

\usepackage{arxiv}

\newcommand{\sysname}{\texttt{SLING}}

\usepackage[utf8]{inputenc} 
\usepackage[T1]{fontenc}    
\usepackage{hyperref}       
\usepackage{url}            
\usepackage{booktabs}       
\usepackage{amsfonts}       
\usepackage{nicefrac}       
\usepackage{microtype}      
\usepackage{lipsum}
\usepackage{graphicx}
\graphicspath{ {./images/} }

\title{Leveraging Semantic Parsing for \\ Relation Linking over Knowledge Bases}

\author{
 Nandana Mihindukulasooriya, Gaetano Rossiello, Pavan Kapanipathi, Ibrahim Abdelaziz, \\ Srinivas Ravishankar, Mo Yu, Alfio Gliozzo, Salim Roukos, Alexander Gray \\
IBM Research, T.J. Watson Research Center,\\ Yorktown Heights, NY, USA\\
\texttt\{nandana.m, Gaetano.Rossiello, ibrahim.abdelaziz1, srini, alexander.gray\}@ibm.com \\
\texttt\{kapanipa, yum, gliozzo, roukos\}@us.ibm.com
}

\begin{document}
\maketitle
\begin{abstract}
Knowledge base question answering systems are heavily  dependent on relation extraction and linking modules. However, the task of extracting and linking relations from text to knowledge bases faces two primary challenges; the ambiguity of natural language and lack of training data. 
To overcome these challenges, we present {\sysname}, a relation linking framework 
which leverages semantic parsing using Abstract Meaning Representation (AMR) and distant supervision. {\sysname} integrates multiple approaches that capture complementary signals such as linguistic cues, rich semantic
representation, and information from the knowledge base.
The experiments on relation linking using three KBQA datasets, QALD-7, QALD-9, and LC-QuAD 1.0 demonstrate that the proposed approach achieves state-of-the-art performance on all benchmarks.
\end{abstract}


\section{Introduction}
Relationship Extraction and Linking (REL) is a necessary task for Knowledge Base Question Answering (KBQA)~\cite{lcquad,DBLP:conf/semweb/UsbeckGN018,qald7}. The goal of REL in KBQA is to identify the relations in input natural language questions and link them to their equivalent relations in a knowledge base, which are then used to construct the corresponding SPARQL query to retrieve answers. For example, we show below  the corresponding DBpedia \cite{lehmann2015dbpedia} SPARQL query for the question ``Who is starring in Spanish movies produced by Benicio del Toro?": 

{\small
\begin{verbatim}
    SELECT DISTINCT ?result WHERE { 
        ?film dbo:starring ?result .
        ?film dbo:country dbr:Spain .
        ?film dbo:producer dbr:Benicio_del_Toro .
        }
\end{verbatim}
}

\noindent Identifying the relevant relations in the question and linking them to their equivalent DBpedia relationships \texttt{dbo:starring}, \texttt{dbo:country}, and \texttt{dbo:producer} is the primary goal of REL in the context of KBQA. 

REL for KBQA faces the following challenges: (1) Knowledge bases such as DBpedia, Wikidata$^{\tiny{\textregistered}}$, and Freebase have a large number of relationships which makes it challenging to acquire training data to build data-intensive deep learning models. For instance, DBpedia has thousands of  relationships (some of which are generated automatically from \textit{Wikipedia}$^{\tiny{\textregistered}}$ infobox keys). (2) There is an extensive lexical gap between the surface form of relations in text and how they are represented in the KB, which makes the linking between them challenging. For example, the question above does not explicitly mention any reference to the relationship \texttt{dbo:country} which is a required relation to form the SPARQL query that can retrieve the answer. (3) Determining multiple relationships and their source and target concepts in a sentence. The example question above requires three relationships to be linked with their corresponding source and target entities/unbound variables. 

In order to address the aforementioned challenges, in this work, we propose our \texttt{Semantic LINkinG} system: {\sysname}; a distant supervision based approach that leverages semantic parsing such as Abstract Meaning Representation (AMR) for relation extraction and linking. Distant supervision techniques address the challenge of lack of training data, particularly for thousands of relations in KBs such as DBpedia. 

Transforming the text to a semantic parse such as AMR, provides advantages that include (1) normalising relations to a set of standard \texttt{PropBank} predicates, (2)  identification of named entities, and  (3) entity typing with a predefined type system. These characteristics of AMR help to alleviate the lexical gap by reducing different phrasings of relations to its predicate set. Furthermore, they also help to automatically determine the relationship structure of an input question and extract all relationships useful for forming a SPARQL query, hence addressing the challenge of extracting multiple relationships from questions text.

In summary, the main contributions of this paper are as follows:
\begin{itemize}
    \item A generic framework integrating different approaches for REL based on statistical predicate alignment, word embedding and neural networks. Furthermore, the framework is modular to allow for integrating more techniques to the pipeline. 
    \item A novel approach that harnesses AMR semantic parses of texts for REL in KBQA. Our novel usage of AMR successfully addresses the lexical gap and multiple relationship problems in REL, and achieves the new state-of-the-art on multiple benchmarks (QALD \cite{DBLP:conf/semweb/UsbeckGN018,qald7} and LC-QuAD 1.0 \cite{lcquad}).

    \item A distant supervised technique that can generate mappings between text, AMR, and KB relations leveraged for training relation classification models in the absence of task-specific training data.

\end{itemize}

The rest of the paper is organised as follows: In Section~\ref{sec:related}, we position our work compared to related work in REL, and Section~\ref{sec:approach} provides an overview of the proposed approach including a summary of each sub-module. In Section~\ref{sec:question_metadata} we describe the metadata generated from the question to be used by the relation linking modules. 
Section~\ref{sec:dist_supervised} describes how the distant supervision data is generated and the two relation linking modules that leverage these data to generate relation linking candidates.
Section~\ref{sec:unsupervised_rel_linking} describes how \sysname\ aggregates the scores from the different relation linking modules and identify the top-k relations.
Next, we evaluate our system in comparison with state-of-the-art relation extraction linking approaches in Section~\ref{sec:results}. Finally, we conclude and present our future work in Section~\ref{sec:conclusion}.

\section{Related Work}
\label{sec:related}

Several KBQA systems have been proposed in the literature which differ according to the traits of the datasets used, such as the amount of training data available, the complexity of the questions, or if the formal queries are provided as ground truth~\cite{DBLP:journals/corr/abs-1907-09361,DBLP:journals/semweb/HoffnerWMULN17}.
In most KBQA systems, extracting relations from the questions and linking them to the KB is an essential step to generate the structure of the formal queries.

REL has been addressed using deep learning models for KBQA datasets with large number of training examples. These deep learning approaches fall into two main categories: classification-based models~\cite{DBLP:conf/semweb/LukovnikovF019} 
and ranking-based models~\cite{DBLP:conf/acl/YuYHSXZ17}.
However, there are drawbacks using end to end neural approaches for REL linking: (1) they are limited only to the questions expressing one single relation in the KB. (2) they  cannot be applied in the case of a lack of training data.

\textsc{QALD}~\cite{DBLP:conf/semweb/UsbeckGN018,qald7} and \textsc{LC-QuAD}~\cite{lcquad}, are well-known datasets derived from DBpedia~\cite{auer2007dbpedia}, represent a real-world evaluation benchmark in evaluating KBQA systems.
The limited amount of training examples of these datasets along with the complex questions involving an arbitrary number of relation types make the task of identifying and linking relations significantly challenging. For KBQA, this is addressed either as a part of an end-to-end question answering system such as 
  \textsc{GAnswer}~\cite{DBLP:journals/tkde/Hu0YWZ18} or by training a model to select the best off the shelf REL system \textsc{Frankenstein}~\cite{DBLP:conf/esws/SinghBRS18}. Our work focuses on building such off the shelf tools, particularly for KBQA.
  
There are four primary REL works that are geared towards the above mentioned KBQA datasets~\cite{earl,rematch,sakor2019old,sibkb}.
\textsc{ReMatch}~\cite{rematch} models every KB relation into a data structure that encapsulates the relation and some enhanced attributes from dependency parsers and WordNet taxonomy. It then applies a number of similarity measures between the question and the KB relations to output a list of candidate relations. \textsc{EARL}~\cite{earl} jointly links relation and entities from natural language to KGs. It extracts the keywords from the question, identifies them as entity or relation, then gets a list of candidates from the KG. Similarly, \textsc{Falcon}~\cite{sakor2019old} is an approach that jointly links entities and relations in question-like sentences to DBpedia. 
For a given question, it applies a number of steps to extract candidate entities and relations including POS tagging, tokenization, compounding and n-gram tiling. Falcon is the state-of-the-art approach for REL on the QALD-7 and LC-QuAD 1.0 datasets. Entity Enabled Relation Linking (EERL)~\cite{pan2019entity} introduces entity-based relation expansion to the existing commonly used keyword based relation extraction with the hypothesis that relations that occur in the question should be either properties of the entities in question or of their types.

However, none of the above mentioned REL methods for KBQA have explored the use of semantic parsers, whilst our work is the first to leverage the AMR of the question text as one of the inputs in an effort to reduce the ambiguity of natural language. Furthermore, we train a distantly supervised neural model in order to address the lexical gap issue between the relations expressed in the questions and the relation labels in the KB. 
This is inspired by the use of distant supervision for standard relation extraction tasks when there is limited or no training data for the target relations~\cite{DBLP:conf/naacl/RossielloGFFG19}.

\section{System Overview}
\label{sec:approach}
\begin{figure}[!h]
    \centering
    \includegraphics[width=1\textwidth]{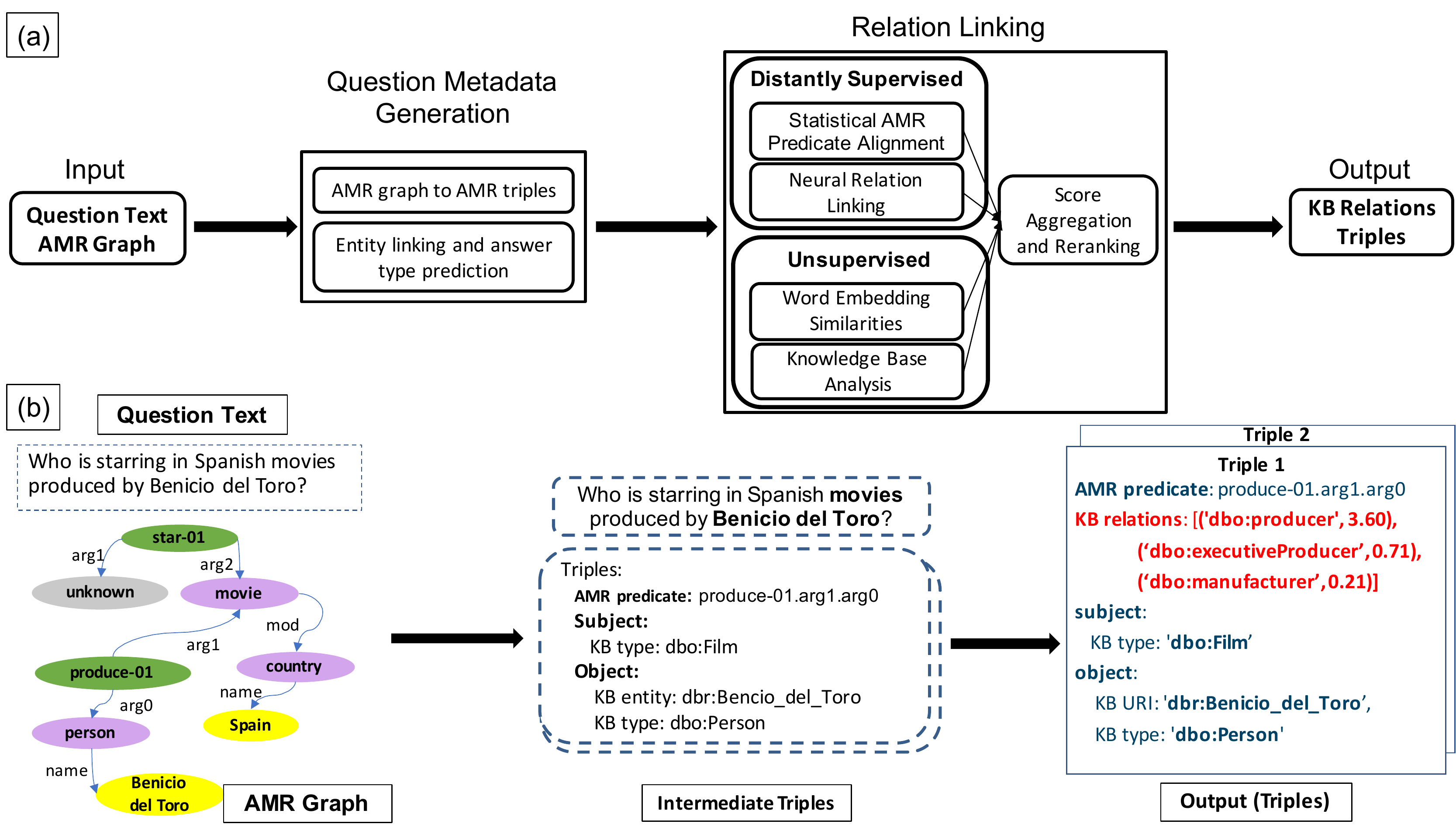}
    \caption{(a) Overview of \sysname. (b) Example driven flow of the approach.}
    \label{fig:arch_overview}
\end{figure}

Figure~\ref{fig:arch_overview} shows an overview of {\sysname} with \ref{fig:arch_overview}-(a) showing a process-oriented view while \ref{fig:arch_overview}-(b) illustrating with an example.
The input to {\sysname} is a question in natural language along with its corresponding AMR representation. The required output is a ranked list of relations corresponding to every subject-object pair in the sentence. The input is processed by the components in \textit{Question Metadata Generation} (Section~\ref{sec:question_metadata}) to extract AMR triples (subject-object pair and their AMR predicate) and generate metadata corresponding to each of them. Each module in \textit{Relationship Linking} produces a ranked list of KB relations with scores for a metadata-enriched AMR triple. These are aggregated to produce the required output. The source code is available at GitHub\footnote{\url{https://github.com/IBM/kbqa-relation-linking}} under Apache 2.0 license.

{\sysname}'s design is modular to allow different relation linking modules to be plugged in and used as needed. The motivation for using multiple modules is to capture different signals such as linguistic cues from the question, richer semantic information from the AMR predicates and roles, semantic similarities of terms, and heuristics from the KB itself.

We have implemented four different relation linking modules. The first two are novel relation linking approaches; both rely on distantly supervised data which we create automatically using the DBpedia and Wikipedia documents (see Section~\ref{sec:dist_supervised}). The other two relation linking modules are unsupervised (see  Section~\ref{sec:unsupervised_rel_linking}). Each of the four modules provides relations with corresponding scores. We aggregate these scores to output a final ranked list of relations. 

An example of the metadata and the output is shown in Figure~\ref{fig:arch_overview} (b). The input data includes the question text and its AMR graph. The modules in \textit{Question Metadata Generation} convert the AMR graph into a set of intermediate AMR triples. Subjects and objects can be either \textit{named entities} such as ``\texttt{Benicio del Toro}'' or \textit{nominal entities} such as ``\texttt{movie}'' (referring a set of unknown movies). Named entities are linked to KB entities and nominal entities to KB classes. This information is passed to individual relation linking modules. Finally, the system generates a set of output triples with a scored ranked list of KB relations. 

\section{Question Metadata Generation}
\label{sec:question_metadata}
The components in \textit{Question Metadata Generation}, process  the question text and its AMR to produce the necessary metadata for relation linking components. The metadata include: (a) AMR triples, (b) KB entities and their types, and (c) answer type prediction. 

\sloppypar
\paragraph{\textbf{AMR Graphs}} As an input, {\sysname} expects a richer semantic representation of the question  generated by an AMR parser~\cite{naseem2019rewarding}. An AMR parse is a rooted, directed, acyclic graph expressing \texttt{"who is doing what to whom"} in a sentence or a question. Figure~\ref{fig:arch_overview}(b) shows a simplified version of an AMR graph for the question \texttt{Who is starring in Spanish movies produced by Benicio del Toro?}. Each node in the graph represents a concept, whereas edges represent relations between concepts that include ProbBank frames, nominal entities (types) and named entities. In this work, we rely on AMR graphs for the following reasons: (1) AMR detects named entities and maps them to predefined entity types (normalized) which forms the arguments of relations that have to be mapped to a KB, (2) AMR not only identifies relations in text but also normalises them using \textit{PropBank} frames; (3) It reduces the ambiguity of natural language by converting relation phrases to their corresponding sense and (4) for questions, a special node, \textit{amr-unknown}, is used to represent a placeholder for the answer to the question. Furthermore, the root node of the AMR graph, a.k.a the focus node, identifies the main focus of the question. Therefore, by using semantic parsing, we abstract out the syntactic variations and capture the meaning of the question in a more normalised manner.

\paragraph{\textbf{AMR graph to AMR triples.}} DBpedia has only binary relations (two arguments). However, frames in AMR can have more than two arguments. For example, the \texttt{produce-01}\footnote{\url{http://verbs.colorado.edu/propbank/framesets-english-aliases/produce.html}} frame can have four  core roles; \texttt{creator} (arg0), \texttt{creation} (arg1), \texttt{created from} (arg2), and \texttt{benefactive} (arg3) and other non-core roles such as time or location whereas on DBpedia there are only binary relations such as \texttt{dbo:producer}, \texttt{dbp:productionDate}, \texttt{dbo:basedOn}, or \texttt{dbo:location}. Despite the richer representation, this inherent mismatch between \textit{n-ary} arguments of PropBank~\cite{kingsbury2003propbank} frames and \textit{binary} predicates in the KB poses a challenge.
 Therefore, it is necessary to generate AMR triples with a similar structure to KB triples (subject, predicate, and object) to facilitate their alignment. 
To resolve this issue, we use an approach that performs combinatorial expansion of all arguments of a frame to create binary relations and then prunes less probable combinations. More details of this process are presented in Section~\ref{sub.sec.amr.alignment}. 

\paragraph{\textbf{Entity/type linking and answer type prediction.}} 
Once the AMR triples are derived, the next step is to link its subject and object to the KB. Subjects and objects from AMR can either be entities (Fig.~\ref{fig:arch_overview}: Bencio del Toro $\rightarrow$ \texttt{dbr:Bencio\_del\_Toro}) or classes (Fig.~\ref{fig:arch_overview}: movie $\rightarrow$ \texttt{dbo:Film}) in the KB. 
Entities are first linked to the KB using a regular entity linking tool that is based on BLINK~\cite{wu2019zero} and DBpedia Lookup. 

For classes, the mapping between AMR type system and DBpedia type system are generated semi-automatically. First, for each of 126 types from AMR type system (from AMR spec\footnote{\url{https://amr.isi.edu/doc/ne-types.html}}), their instances are collected from AMR graphs and linked to KB entities. Then KB entity types are collected and they are ranked by frequency. Top 5 types are checked manually to map a KB type to each AMR type. This is a one time process that takes $\sim$ 2 hours. This mapping can be performed against any type system (e.g., DBpedia, Wikidata$^{\tiny{\textregistered}}$) given a tool for entity linking is available. For the special node, \texttt{amr-unknown}, we map it to a KB-type by using an LSTM-based answer type prediction model. For instance, given a question such as ``\texttt{Who is starring in Spanish movies produced by Benicio del Toro?}'', it predicts \texttt{dbo:Actor} as the answer type.

\section{Distantly Supervised Relation Linking}
\label{sec:dist_supervised}
The question metadata such as AMR parse, AMR triples with entity and type information (from Section~\ref{sec:question_metadata}) are used as input to the four REL modules. Two of the modules that rely on distant supervision data are described below and the other two in the next Section~\ref{sec:unsupervised_rel_linking}. 

Distantly supervised data is generally used in tasks where there is a lack of training data~\cite{DBLP:conf/acl/MintzBSJ09}. The lack of training data is also a significant challenge for REL tasks on KBQA datasets. Particularly, if we want to perform REL to DBpedia, we need training data for thousands of DBpedia relations. On the other hand, the KBQA datasets such as QALD and LC-QuAD 1.0 have 408 and 5000 questions covering a small subset of DBpedia relations. In order to address this issue, we collect training data using distant supervision, which eliminates the need for task-specific supervision for relation linking. 


\subsection{Distant Supervision Dataset}
\label{sec:dataset}
\begin{figure}[!ht]
    \centering
    \includegraphics[width=1\textwidth]{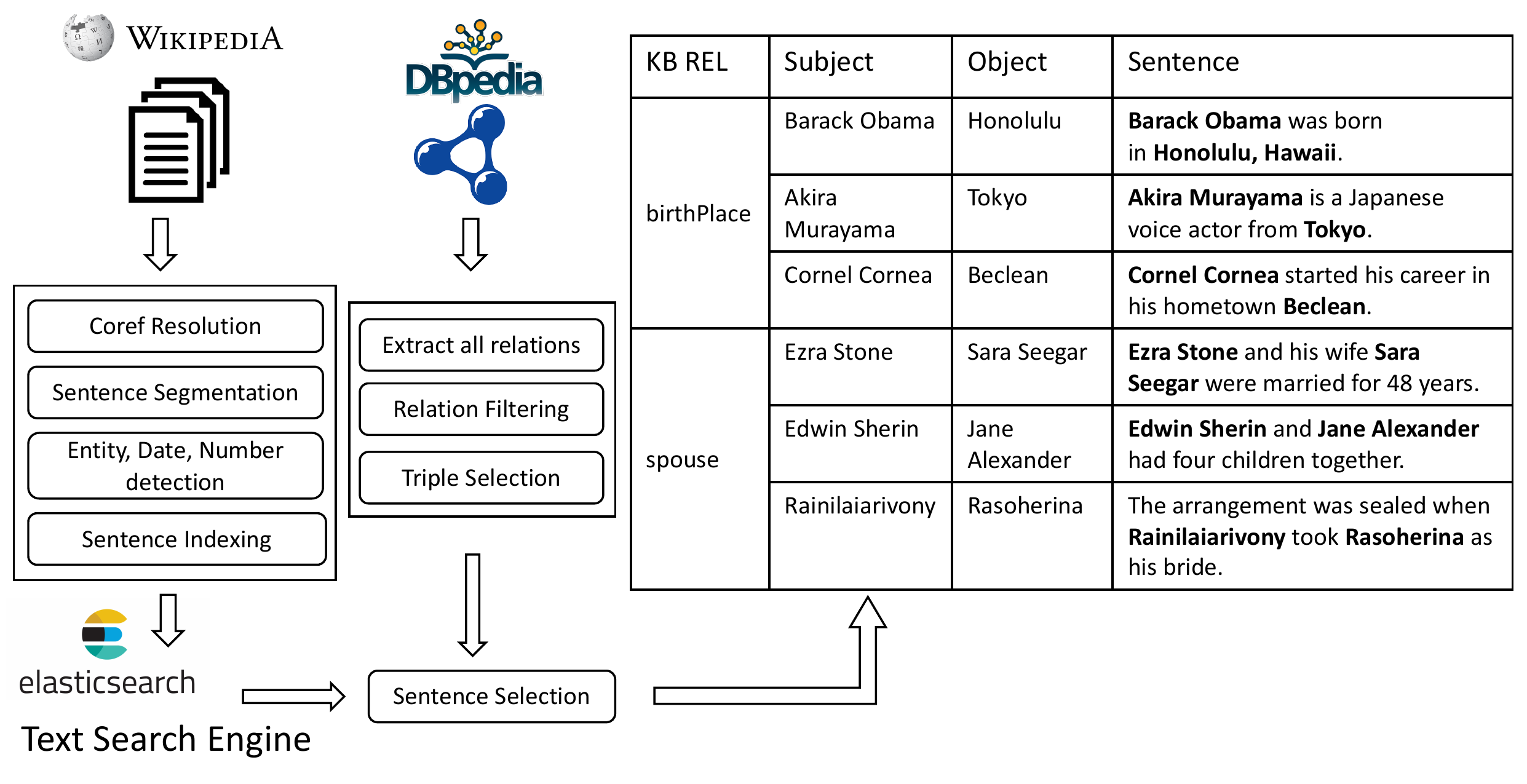}
    \caption{Distant supervision data generation pipeline}
    \label{fig:distant_supervision}
\end{figure}

To train our REL models, for each relation, we require training examples (sentence, subject, object) mapped to its corresponding KB relation. For instance, as shown in Figure~\ref{fig:distant_supervision} (Sentence: \texttt{Barack Obama was born in Honolulu, Hawaii}, subject: \texttt{Barack Obama}, object: \texttt{Honolulu, Hawaii}) mapped to (KB relation: \texttt{dbo:birthPlace}). 


{\textbf{Corpus pre-processing and indexing:}} As shown in Figure \ref{fig:distant_supervision}, we begin with the Wikipedia$^{\tiny{\textregistered}}$ corpus, and perform co-reference resolution on each document. The corpus is then tokenized into sentences, and named entities are identified in each sentence to serve as \texttt{ElasticSearch} indices. We also store meta-data such as the document the sentence was extracted from and its position in the document. This meta-data is later used for selecting sentences.

{\textbf{Relation selection:}} To address the issue of the large number of relations in KB, we select a manageable subset. DBpedia has a long tail of relations mainly due to uncommon Wikipedia$^{\tiny{\textregistered}}$ infobox keys that are not widely used in queries. The number of examples that are generated by the distant supervision process depends on the number of triples containing the relation in the KB. While unsupervised modules use all relations in DBpedia, distantly supervised modules require some amount of examples to train the modules; thus the number of relations used by them depends of their frequency of occurrence. The distance supervision process can generate more than 10 examples for $\sim$ 1.3K relations.


{\textbf{Selection of examples:}} For each relation, we pick up to 1000 KB triples by ordering them by the sum of subject and object in-degrees. The assumption is that these entities are central and generally their corresponding Wikipedia$^{\tiny{\textregistered}}$ articles contain more information. Then for each KB triple, we select a single example sentence, which is the first cooccurrence in entity's Wikipedia$^{\tiny{\textregistered}}$ article. We choose sentences that satisfy the following: 1) subject and object co-occur, 2) have at least 4 tokens, 3) have at least 1 verb and 4) the entity surface forms do not overlap in the text (when one is a multi-word containing other). We observed that these basic heuristics increased the probability that the sentence contains a relation and filtered out accidental co-occurrences in titles, lists, etc. 


\subsection{Statistical AMR Predicate Alignment}
\label{sub.sec.amr.alignment}
This section presents a relation linking module that leverages the information present in the AMR semantic parses to generates alignments between \textit{PropBank} predicates in AMR graphs and KB ontology relations. We describe below how these alignments are generated and then used to produce candidate relations. 

\subsubsection{Building PropBank Alignments}  
\begin{figure}[h]
    \centering
    \includegraphics[width=1\textwidth]{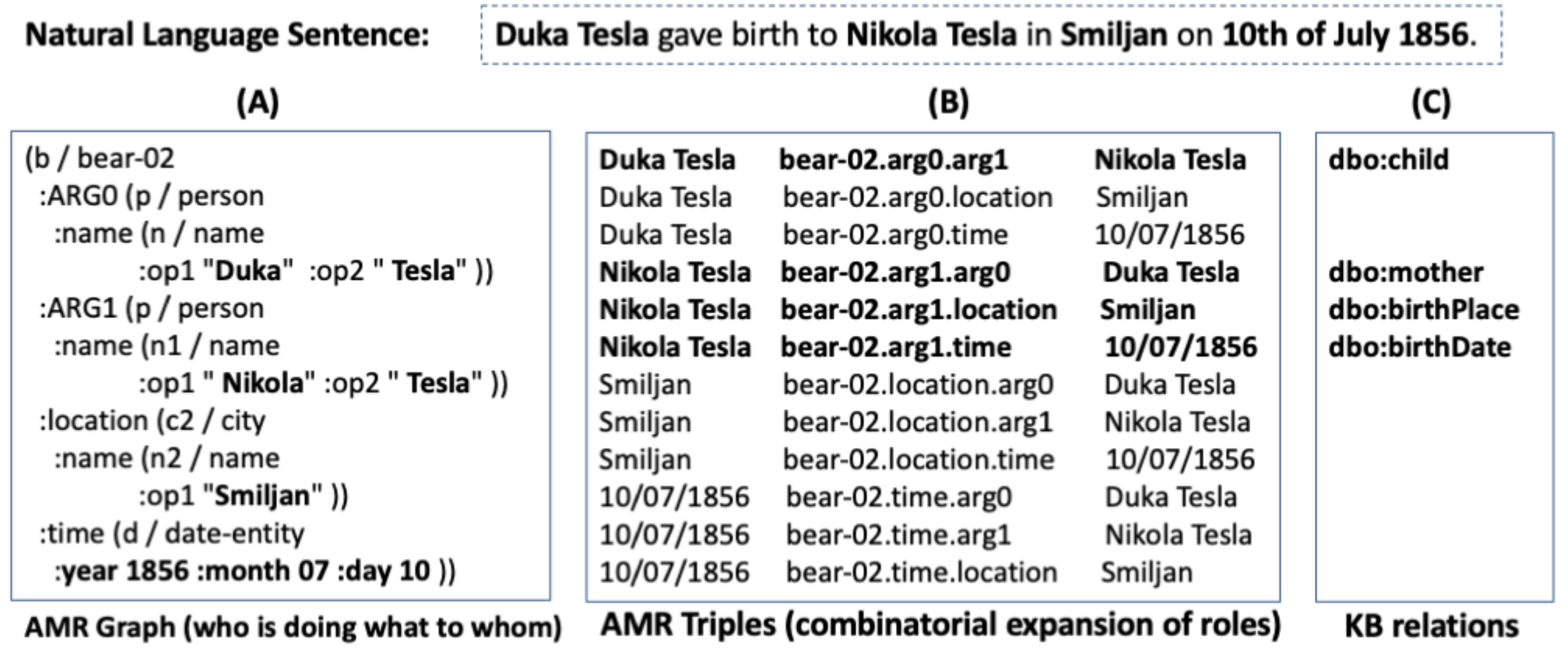}
    \caption{converting AMR graphs to binary relations using combinatorial expansion}
    \label{fig:amr_to_triples}
\end{figure}

One challenge for creating these alignments is the inherent mismatch between frame-based representation and triple-based representation. In AMR graphs, a single frame captures a rich set of information using \textit{n-ary} relations (e.g., who is doing what to whom, when, etc.) while triples in KBs capture simpler atomic facts using \textit{binary} relations. 
For example, the frame \texttt{bear-02}\footnote{\url{http://verbs.colorado.edu/propbank/framesets-english-aliases/bear.html}}, which is used to capture the event of giving birth to a child, has two core roles: \texttt{arg0} (mother), \texttt{arg1} (child) and several non-core roles including \texttt{location} (place of birth), \texttt{time} (time of birth) as shown in Figure~\ref{fig:amr_to_triples}-A.

To address this mismatch in the number of arguments, we first decompose the AMR graph into a set of AMR triples. This is performed by creating binary relations between all entities participating in different roles of the frame using combinatorial expansion, as shown in Fig.~\ref{fig:amr_to_triples}-B. The generated binary relations are paths between the two nodes in the graph and follow the structure, \textless propbank-frame\textgreater.\textless subject-role\textgreater.\textless object-role\textgreater. Given a combination of two entities, for example, \textit{Duka Tesla} (with the arg0 role) and \textit{Nikola Tesla} (with the arg1 role), two AMR triples are generated, one with Dula Tesla as subject and Nikola Tesla as object and the other vice-versa as shown below:
{\small
\begin{verbatim}
    Duka Tesla      bear02.arg0.arg1    Nikola Tesla
    Nikola Tesla    bear02.arg1.arg0    Duka Tesla
\end{verbatim}
}

Nevertheless, this process generates a large number of AMR triples that will not necessarily have their mapping relation in the KB. For example, in DBpedia, the place or the date that a mother gave a birth to a child (bear02.arg0.location/time) is not represented as an attribute of the mother but only as attributes of the child and consequently there is no equivalent relation for those in the KB.
This can be addressed by analysing how often we can align a given AMR triple to a KB triple. For example, out of 12 AMR triples generated (Fig.~\ref{fig:amr_to_triples}-B), only the four highlighted can be aligned with the existing KB triples in DBpedia.

Because KBs are generally multi-graphs and there are cases where two entities are connected with multiple relations in the KB. For example, if we assume Nikola Tesla was born and died in the same place, two entities (Nicola Tesla and Smijan) will be related both by \texttt{birthPlace} and \texttt{deathPlace} relations. In such cases with multiple candidates, we use lexical similarity between frame definition/aliases from PropBank (e.g., \texttt{bear}, \texttt{bear children}, \texttt{birth}, \texttt{give birth}) and DBpedia relation labels to disambiguate and select the most similar one. 

\begin{figure}[h]
    \centering
    \includegraphics[width=1\textwidth]{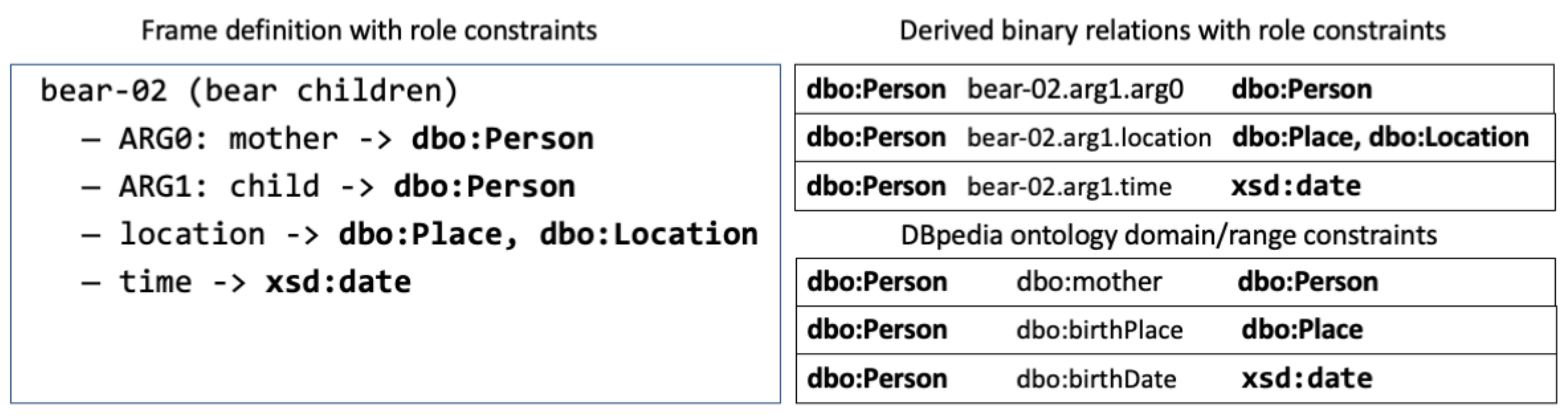}
    \caption{Type constraints for frame roles}
    \label{fig:type_constraint}
\end{figure}

Finally, to accommodate error propagation from both distant supervision dataset and AMR parsing, which could lead to noise in the alignments, we also use type constraints to further refine the alignments. The goal of this step is to induce type constraints for each role in a given frame. This is performed by collecting all entities participating in a given role in a frame (\textit{e.g.}, bear-02.ARG0) and analyzing their types (including data types such as numerics and dates). Using this information, proxy domain and range constraints for AMR binary relations can be generated as in Fig.~\ref{fig:type_constraint}. These constraints are used to filter out any aligned DBpedia relation that does not match with the type constraints. 

To summarize, for generating these alignments efficiently, we used the distant supervision dataset, defined as $D = \{(s_{i}, r_{i}, o_{i}, t_{i}),\dots\}$ where \{$s_{i}, r_{i}, o_{i}$\} are the subject, relation, object of the KB triple and $t_{i}$ is the corresponding sentence (see Fig.~\ref{fig:distant_supervision}). We parse each $t_{i}$ and generate an AMR graph $a_{i}$. Each $a_{i}$ is then converted into a set of AMR triples  $x_{j} = \{\grave{s}_{j}, p_{j}, \grave{o}_{j}\}$ where $x_{j} \in a_{i}$ and $p_{j}$ is the AMR binary predicate, $\grave{s}_{j}$ and $\grave{o}_{j}$ are the subject and object from the AMR graph. 
Finally, we check for an AMR triple $x_{j}$ where $\grave{s}_{j} = s_{i}$ and $\grave{o}_{j} = o_{i}$ and if found, one alignment between $r_{i}$ and $p_{j}$ is created. 

\subsubsection{Finding relation candidates} 
Once the complete dataset is processed and alignments are filtered using type constraints, for each AMR binary predicate $p_{j}$ we get a set of cumulative alignments $A(p_{j}) = \{(r_{0}, c_{0}), \dots, (r_{n}, c_{n})\}$ where each $r$ is a KB relation and $c$ is a alignment count. Using that, for each AMR binary predicate $p_{j}$, relation candidate scores are calculated using $relation\_score(p_{j},r_{n}) = [c_{n} / max(c)] * [1 / 1 + log(inv\_pred\_count(r_{n})] $ where $max(c)$ is highest count in $A(p_{j})$ and $inv\_pred\_count(r_{n})$ is the inverse predicate count, \textit{i.e.}, number of distinct AMR predicates which $r_{n}$ is aligned at least once.

\subsection{Neural Model for Relation Linking}
\label{neural_model}

Statistical AMR mapping has the following drawbacks that can be addressed using a neural approach: (1) mapping generic frames such as \texttt{have-01} can be ambiguous. For example for: \texttt{``Did Che Guevara have children?"} has the frame \texttt{have} that needs to be mapped to \texttt{dbo:child} (2) lexical gap where 
the same relation type can be expressed as different linguistic patterns.
Therefore, we train a neural model for relation linking by exploiting the distant supervision dataset (Section \ref{sec:dataset}). The neural model produces dense embedding vectors for input questions, which can learn to project the same relation type's different surface forms close in the latent space.
In this section, we describe how to train the neural model on the distant supervision data (\emph{training}) and how to make use of the model for question REL (\emph{inference}).

\subsubsection{Training phase}
Leveraging our distant supervision dataset, our training data is defined as $D = \{(x_{0}, r_{0}), \dots, (x_{N}, r_{N})\}$. Here $r_{i} \in R$ are the relation types, and $x_{i} = (t_{i}, s_{i}, o_{i})$ are the relation instances consisting of a textual sentence $t_{i}$ and the spans of the subject $s_{i}$ and the object $o_{i}$.
The set $R$ represents the vocabulary of $K = |R|$ distinct relation types.
We train a neural network $M$ on $D$ with the purpose to predict the correct relation type $r_{k}$ given the instance $x_k$ by minimize the cross-entropy loss regarding the conditional probability $p_{M}(\cdot|x_k)$ modeled by $M$, with respect to the true relation $r_k$.

In order to generate a vector representation of the relation instance $x$, we adopt the relation encoder inspired by \cite{DBLP:conf/acl/SoaresFLK19}. This encoder is an adaptation of the original Transformer \cite{DBLP:conf/nips/VaswaniSPUJGKP17} architecture that encodes the given sentence while being aware of the subject and object. 
To achieve this entity-aware encoding, we introduce four special tokens to mark the start and end positions of both entities in the sentence, \texttt{[SUBJ]}, \texttt{[\textbackslash SUBJ]}, \texttt{[OBJ]} and \texttt{[\textbackslash OBJ]} respectively.
For instance, the second relation instance of \texttt{birthPlace} in Fig \ref{fig:distant_supervision} is represented as ``\texttt{[SUBJ] Akira Murayama [\textbackslash SUBJ] is a Japanese voice actor from [OBJ] Tokyo [\textbackslash OBJ]}''.
These new special tokens are randomly initialized and fine-tuned during training, whereas all the other tokens are initialized using the pre-trained \textsc{BERT-base} embeddings~\cite{DBLP:conf/naacl/DevlinCLT19}.
We concatenate the vectors of the final-layer hidden states of the start entity markers of subject and object entities, feed them into a fully connected layer to get the finally embedding vector for the relation instance $x$.

Finally, to estimate $p_M(r_{k}|x_k)$, we add a further classification layer with the output size $K$ followed by a softmax function. 

\subsubsection{Inference phase}
There are several challenges to address when applying the trained neural relation linking model $M$ to deal with question relationship linking. In particular, how to mark the missing entities from the question, which consists of two cases: (1) the missing entity is the answer; (2) the missing entity is an intermediate entity when the question requires multiple hops to reach the answer.
We exploit the \textit{AMR graph to AMR triples} feature described in Section \ref{sec:question_metadata} to handle these challenges.

$\bullet$ \emph{Intermediate entities:}
Consider the question in Fig. \ref{fig:arch_overview} and its generated metadata. 
The question requires to first find some Spanish movie entities having the \texttt{dbo:producer} relation with \texttt{Benicio del Toro}, e.g., \texttt{7 días en La Habana}, then identifies another relation \texttt{dbo:star} from the movie entity. Since the movie name is missing in the question text, when predicting its relationship to \texttt{Benicio del Toro}, we take the surface form of \texttt{arg2} for \texttt{star-01.arg2.arg1}, i.e., the word ``\texttt{movies}'' as the object.
In this way, we generate the following input relation instance to the neural model $M$, \texttt{Who is starring in Spanish [OBJ] movies [\textbackslash OBJ] produced by [SUBJ] Benicio del Toro [\textbackslash SUBJ] ?}. 

$\bullet$ \emph{Unknown (answer) entities:}
Consider the same question as above. The predicate \texttt{star-01.arg2.arg1} has no explicit text for the \texttt{arg1} since the \texttt{amr-type} is \texttt{unknown}, which refers to the answer. 
In this case, we mark the question word ``Who'' for the \texttt{arg2}. 
Therefore the following format for the relation instance is generated for our neural model: ``\texttt{[OBJ] Who [\textbackslash OBJ] is starring in Spanish [SUBJ] movies [\textbackslash SUBJ] produced by Benicio del Toro?}''. 

Finally, with the aforementioned treatments, for each relation instance a ranked list of relation types in DBpedia is generated and sorted by their probability scores produced by our neural model.

\section{Unsupervised Relation Linking and Score Aggregation}
\label{sec:unsupervised_rel_linking}


In the previous section, we have described the 2 distantly supervised modules. In this section, we describe the remaining 2 modules and aggregation of scores to get the final ranked list of KB relations.


\subsection{Unsupervised modules}

\subsubsection{Lexical Similarity}
To derive the score of a relation with respect to an AMR triple, we compute its lexical similarity to the question text and AMR predicate. For each relation candidate, like \texttt{dbo:deathPlace}, we consider its label as a word sequence \texttt{death place}. We concatenate each question, e.g., \texttt{Who was married to Lincoln}, with the AMR predicate of the triple (eg. \texttt{marry} from \texttt{marry-01}) to get the other word sequence.
We compute the lexical similarity between the two word sequences by first calculating a word-by-word cosine similarity based on word2vec embeddings.
If there are $m$ words in one word sequence, and $n$ words in the other, this produces $m \times n$ similarity scores. This is max-pooled to produce a single score as output.

\subsubsection{Knowledge Base Connections}
In KBQA, the entities from the questions are identified and linked to KB first.
Therefore, the task of relation linking also assumes the existence of such linked KB entities and entity types as described in the \textit{Question Metadata Generation} step. 
Hence, candidate relations that also connect these detected entities can be scored higher, following previous works~\cite{sakor2019old}.
For example, given the question ``\texttt{Who created Family Guy?}", to predict the relation in this question, we score all relations connected to the KB entity (\texttt{dbr:Family\_Guy}) as the object and a subject of KB type \texttt{dbo:Person} or any subclass of it (which is predicted by answer type prediction). 
We then apply a soft constraint to focus more on the relations that are within this set.\footnote{Ideally, we can do the hard filtering with the relation connections. However as REL is a component of a whole KBQA pipeline. To mitigate potential error propagation from entity linking, most works adopt a soft approach~\cite{sakor2019old,DBLP:journals/tkde/Hu0YWZ18}.}.


\subsection{Score aggregation}  
The scores from each module are normalized using min-max normalization.
The final score of a relation is the arithmetic sum of its normalized score from each module, and a ranked list of relations is obtained for the AMR triple. This process is repeated for every AMR triple extracted from the question.

\section{Evaluation}
\label{sec:results}
In this section, we detail our experimental setup and evaluate our approach against the state-of-the-art relation linking approaches for KBQA. We replicate the experimental setup proposed in Falcon~\cite{sakor2019old} in terms of the same datasets and metrics used for a fair comparison, as described below.

\begingroup
\setlength{\tabcolsep}{4pt}
\renewcommand{\arraystretch}{1}
\begin{table}[h]
\centering
\small
\begin{tabular}{lcc}
\toprule
Dataset & Questions & Avg Constraints \\
\midrule
QALD-7 & 215 & 1.5\\
QALD-9 & 408 & 1.5\\
LC-QuAD 1.0 & 5000 & 1.7\\
\bottomrule
\end{tabular}
\caption{KBQA datasets statistics}
\label{tab:dataset_deetails}
\end{table}
\endgroup

\subsection{Experimental Setup}


    We used three KBQA datasets; QALD-7~\cite{qald7}, QALD-9~\cite{DBLP:conf/semweb/UsbeckGN018} and LC-QuAD 1.0~\cite{lcquad}. All the datasets comprise of question text, their corresponding SPARQL queries, and answers from DBpedia. Similar to~\cite{sakor2019old}, we use the question text and the relations in the SPARQL queries for evaluation\footnote{We exclude \texttt{rdf:type}, and \texttt{rdfs:label} to follow same setting in \cite{sakor2019old}.}. Table~\ref{tab:dataset_deetails} shows the number of questions and the average triple constraints in SPARQL queries for each of the datasets. QALD-9 is an evolved version of QALD-7 extending the number of questions from 215 to 408.  


We compare \sysname\ against four existing REL approaches for KBQA: (1) SIBKB~\cite{sibkb}, (2) ReMatch~\cite{rematch},  (3) EARL~\cite{earl}, and (4) Falcon~\cite{sakor2019old}. Falcon~\cite{sakor2019old} is the state-of-the-art approach evaluated on QALD-7 and LC-QuAD 1.0 datasets. We use standard metrics such as precision, recall, and F-measure for evaluation and comparisons. The precision measures the capability of a REL system to predict the exact number of expected relations in a given question and the recall measures the capability of a system to cover all the expected relations.


\subsection{Results}

\begingroup
\setlength{\tabcolsep}{6pt}
\renewcommand{\arraystretch}{1}
\begin{table}[t]
\centering
\begin{tabular}{l|ccc||ccc||ccc}
\toprule
 & \multicolumn{3}{c||}{QALD-7}   & \multicolumn{3}{c||}{LC-QuAD 1.0} & \multicolumn{3}{c}{QALD-9}  \\
\midrule
System & P & R & F1 & P & R & F1 & P & R & F1 \\
\midrule
SIBKB   &  0.29 & 0.31 & 0.30 & 0.13 & 0.15 & 0.14 & - & - & -       \\
ReMatch &  0.31 & 0.34 & 0.33 & 0.15 & 0.17 & 0.16 & - & - & -    \\
EARL    &  0.27 & 0.28 & 0.27 & 0.17 & 0.21 & 0.18 & - & - & -   \\
Falcon & \bf0.58 & 0.61 &  0.59 & \bf0.42 & 0.44 & 0.43 & 0.31 & 0.34 & 0.32  \\
\sysname  & 0.57 & \bf0.76 & \bf 0.65 & 0.41 & \bf 0.58 & \bf 0.48 & \bf0.50 & \bf0.64 & \bf0.56 \\
\toprule
\end{tabular}
\caption{Relation Linking systems comparison}
\label{tab:results}
\vspace{-0.2in}
\end{table}
\endgroup

Table~\ref{tab:results} shows the precision, recall, and F-measure of  \sysname\ in comparison to state-of-the-art approaches. The results show that our approach consistently achieves a better F1 score than the existing approaches and is robust across datasets, i.e. the results on QALD-7 and QALD-9 are respectively similar compared to those obtained by Falcon\footnote{Falcon numbers on QALD-7 and LC-QuAD 1.0 are taken from their paper.}.
Moreover, {\sysname} provides a remarkably higher recall than the other competing systems.  


\textbf{Ablation Study: } In order to understand the contribution of each module in the \sysname\  framework, we perform an ablation study by removing the corresponding module from the overall system and comparing its performance. 
These results reported in Table \ref{tab:ablation} indicates that every module contributes to the overall performance of the system results. 
Particularly, removing the statistical AMR mapping approach from the system has the biggest drop in performance. The AMR mapping component provides the strongest contribution of the modules, with the system performance dropping considerably without its usage. 
AMR provides predicates that are already a strong signal for identifying the relations in a sentence. 
Moreover, AMR parsers normalize syntactic variations across sentences which have the same meaning. 
Finally, AMR provides type information about the subject and object of a relation, even when they are unknown. 
This enables domain and range-derived features to constrain the predicted relation candidates. 

For relations that are implicit in text, the Neural Relation Linking bridges the lexical gap to map them. For instance, considering the first example in Table \ref{tab:qualitative}, the relation type \texttt{dbo:country} is implicit in the question, but the neural model is able to identify it nevertheless.
Furthermore, the Neural Relation Linking is able to handle questions having more than one relation, where different relations can be predicted given the same question text, but different spans of entities, as described in Section \ref{neural_model}.
This insight confirms that the distant supervision technique can be helpful in covering different language phrases to identify and link relations to a KB, especially in setting such as QALD, where only a small training set is provided.

Based on our analysis we find that the Word Embeddings module is particularly useful when the relation is explicitly mentioned in the question, like \texttt{`Who is the mayor of Paris?'}, with the relation being \texttt{dbo:mayor}. 
It provides high-precision estimates about the relations in the question. 
\begingroup
\setlength{\tabcolsep}{6pt}
\renewcommand{\arraystretch}{1}
\begin{table}[t]
\centering
\begin{tabular}{lccc}
\toprule
                            & P & R & F1  \\
\midrule
SLING                         & 0.57      & 0.76   & 0.65\\
~~w/o AMR Mapping             & 0.45      & 0.57   & 0.51  \\
~~w/o Neural Relation Linking               & 0.52      & 0.66   & 0.58 \\
~~w/o Word Embeddings    & 0.53      & 0.68   & 0.59  \\
~~w/o KB Analysis             & 0.46      & 0.61   & 0.53 \\
\bottomrule
\end{tabular}
\caption{Ablation study on QALD-7 dataset}
\label{tab:ablation}
\end{table}
\endgroup
\begingroup
\setlength{\tabcolsep}{6pt}
\renewcommand{\arraystretch}{1}
\begin{table}[t]
\centering
\small
\begin{tabular}{lccc}
\toprule
                            & P & R & F1  \\
\midrule
w/ machine generated AMR  &  0.53    &  0.76  & 0.62 \\
w/ human annotated AMR & 0.57  & 0.77 & 0.66 \\
\bottomrule
\end{tabular}
\caption{Relation linking performance with machine generated vs human annotated AMR on a subset of QALD-9 dataset}
\label{tab:amr_comparision}
\end{table}
\endgroup
Automatic extraction of DBpedia triples from Wikipedia Infoboxes (when mappings are not available) introduces redundant and noisy (\texttt{dbp:}) relations. For instance, there are relations such as \texttt{dbo:birthPlace}, \texttt{dbp:birthPlace}, \texttt{dbp:birthLocation} and \texttt{dbp:placeOfBirth} that are semantically equivalent and cannot be lexically distinguished based on their labels. In such scenarios KB analysis allows the system to choose the correct relation by considering only the ones connected to the entities of interest. For the question \texttt{`What is the birth place of Frank Sinatra?'}, without the KB analysis we find \texttt{dbo:birthPlace} and \texttt{dbp:placeOfBirth} as the top ranked relations. KB analysis scores \texttt{dbp:placeOfBirth} higher because of its association with the entity \texttt{dbr:Frank\_Sinatra}.

\textbf{Impact of AMR Parser: } To understand how the quality of AMR affects the results, we have manually annotated a subset of QALD-9 questions (ids 250 to 408) and the results are presented in Table~\ref{tab:amr_comparision}. It shows that human annotated AMRs provide an improvement of 4 points in F1. The state-of-the-art AMR parser~\cite{naseem2019rewarding} has a smatch score of 90\% when tested on a subset of QALD-9 dataset. 

\textbf{Impact of Entity Linking: }To understand the effect of entity linking, we have performed a similar experiment using entity annotations provided by \cite{pan2019entity} for LC-QuAD 1.0. We have tested the entity linker implementation we used with QALD-9; it has an F1 of 0.75.

\begingroup
\setlength{\tabcolsep}{6pt}
\renewcommand{\arraystretch}{1}
\begin{table}[t]
\centering
\small
\begin{tabular}{lccc}
\toprule
                            & P & R & F1  \\
\midrule
w/ our entity linker implementation  &  0.41    &  0.58    & 0.48 \\
w/ entity annotations from \cite{pan2019entity} & 0.46  & 0.62 & 0.53 \\
\bottomrule
\end{tabular}
\caption{Relation linking performance with our entity linking implementation vs annotations from \cite{pan2019entity} on LC-QuAD dataset.}
\label{tab:entity_linking}
\end{table}
\endgroup

\subsection{Qualitative Analysis and Discussion}

Table~\ref{tab:qualitative} shows five example questions with their gold standard relations compared to what \sysname\ predicts for each question. \texttt{SLING} was able to find the correct set of relations for the first three questions and partially solves the rest. In the first question, the main  challenge is to decompose the question into three triples with correct subject/object combinations. Leveraging AMR allows \sysname\ accurately determine the correct triple decomposition including directionality and the number of triples. Once decomposed, all relation linking modules provide strong signals in this example.  The second and third questions are lexically very similar but their representations in KB are different. The fact that \texttt{SLING} creates triples with directionality into account and perform KB analysis allowing it to pick correct relation in each case.  The fourth question is challenging because it requires to reason that grand children are children of children. AMR represent this using a single triple. Furthermore, the relation used in each constraint is different (\texttt{dbo:child} vs \texttt{dbp:child}). \texttt{SLING} gets a set of candidates such as \texttt{dbo:child}, \texttt{dbp:children}, \texttt{dbp:grandChilden} and picks \texttt{dbo:child} as it is connected to \texttt{res:Bruce\_Lee}. Nevertheless, it does not decompose this question into two triples. Similarly, some gold standard questions have the \texttt{UNION} construct with logically equivalent relations in KB. However, as the relation appears only once in the text, \texttt{SLING} only aims to predict one relation. 
 
\begin{table}[h!]
\begin{tabular}{lp{5.5cm}lp{3cm}}
\toprule
ID &
  Question &
  Gold standard triple patterns &
  Predicted relations \\ 
  \midrule
1 &
  Who is starring in Spanish movies produced by Benicio del Toro? &
  \begin{tabular}[c]{@{}l@{}} ?film dbo:starring ?actor .\\ ?film dbo:country res:Spain .\\ ?film dbo:producer res:Benicio\_del\_Toro\end{tabular} &
  \begin{tabular}[c]{@{}l@{}}dbo:starring\\ dbo:country\\ dbo:producer\end{tabular} \\ \midrule
2 &
  Who developed Skype? &
  res:Skype dbo:developer ?company &
  dbo:developer \\ \midrule
3 &
  Who developed Slack? &
  ?company dbo:product  res:Slack &
  dbo:product \\ \midrule
4 &
  Give me the grandchildren of Bruce Lee. &
  \begin{tabular}[c]{@{}l@{}}res:Bruce\_Lee dbo:child ?child\\ ?child  ?dbp:child  ?granchild\end{tabular} &
  dbo:child \\ \midrule
5 &
  Which organizations were founded in 1950? &
  \begin{tabular}[c]{@{}l@{}}\{ ?org dbo:formationYear ?date \} UNION \\ \{ ?org dbo:foundingYear ?date \} UNION\\ \{ ?org dbp:foundation ?date \} UNION \\ \{ ?org dbp:formation ?date \}\end{tabular} &
  dbo:foundingYear \\ 
  \bottomrule
\end{tabular}
\caption{Example queries with gold and predicted relations.}
\label{tab:qualitative}
\end{table}

\section{Conclusions and future work}
\label{sec:conclusion}

In this paper, we presented {\sysname}, a framework for relation linking that leverages semantic parsing with AMR and distant supervision. {\sysname} is a combination of multiple modules that capture complementary signals both from the AMR representation as well as natural language text. Experimental results show that {\sysname} outperforms state-of-the-art approaches on three KBQA datasets; QALD-7, QALD-9, and LC-QuAD 1.0. 
Furthermore, our ablation study shows that leveraging AMR and the use of distant supervision contributes to outperform the state-of-the-art techniques. 
As a part of our future work, we are planning to convert all the components as feature generators for an end to end neural approach. Furthermore, we intend to investigate the use of transformer-based architectures for encoding both AMR graphs and the question text for relation linking.

\bibliographystyle{unsrt}  
\bibliography{ms}

\end{document}